\documentclass[a4paper]{article}
\usepackage{INTERSPEECH2014,amssymb,amsmath,epsfig}
\ninept
\def\reg{{\rm\ooalign{\hfil
     \raise.07ex\hbox{\scriptsize R}\hfil\crcr\mathhexbox20D}}}

\title{Bayesian calibration for forensic evidence reporting}

\makeatletter
\def\name#1{\gdef\@name{#1\\}}
\makeatother \name{{\em Niko Br\"ummer and Albert Swart}}

\address{$^1$AGNITIO Research, South Africa \\
{\small \tt \{nbrummer,aswart\}@agnitio-corp.com}}

\begin{document}
\maketitle
\begin{abstract}
We introduce a Bayesian solution for the problem in forensic speaker recognition, where there may be very little background material for estimating score calibration parameters. We work within the Bayesian paradigm of evidence reporting and develop a principled probabilistic treatment of the problem, which results in a Bayesian likelihood-ratio as the vehicle for reporting weight of evidence. We show in contrast, that reporting a likelihood-ratio distribution does not solve this problem. Our solution is experimentally exercised on a simulated forensic scenario, using NIST SRE'12 scores, which demonstrates a clear advantage for the proposed method compared to the traditional plugin calibration recipe.
\end{abstract}
\noindent{\bf Index Terms}: forensic speaker recognition, Bayesian paradigm

\section{Introduction}
The problem that we address is how to use the output (score) of a speaker recognizer to make  minimum-expected-cost Bayes decisions. Bayes decision theory provides an \emph{idealized mathematical model} for the legal process in a court case, where it has to be established whether the prosecution hypothesis holds \emph{beyond a reasonable doubt}. One interpretation of this model is that it quantifies \emph{doubt} via posterior probabilities, while the \emph{reasonable} threshold can be expressed via the relative costs of false acquittal and false conviction. Specifically, this model prescribes, convict if
\begin{align}
\label{eq:bdt}
\frac{P(\text{prosecution hypothesis}|\text{evidence})}{P(\text{defence hypothesis}|\text{evidence})} > \frac{C_\text{false-convict}}{C_\text{false-acquit}}
\end{align}
The LHS represents the doubt, while the RHS is the reasonable threshold. Another interpretation is that it chooses the course of action (convict or acquit) which minimizes the expected cost of making a wrong decision. 

The threshold could instead be motivated without explicit reference to cost. Jaynes~\cite{book:Jaynes_PTTLOS_2003} suggests: Setting the threshold at 10~000 ``will mean, crudely,
that on the average not more than one conviction in 10~000 will be in error; a judge who required juries to follow this rule would probably not make one false conviction in a working lifetime on the bench''.

The purpose of this paper is not to debate the applicability of speaker recognition or Bayes decision theory to real court cases---for examples of such debate, see~\cite{caution,rose_science_law}. Our goal is to design speaker recognizers to produce outputs with a clearly defined interpretation. This interpretation is supplied by the Bayes decision model. We work within this model and try to make our recognizer as good as possible according to this interpretation.

\section{Theory}
We shall restrict ourselves to the scientific problem of computing the LHS of~\eqref{eq:bdt}, with the assumption that the legal process will take care of quantifying the RHS. We follow Balding's principle~\cite{balding2005weight}: ``Evidence is of value inasmuch as it alters the
probability that the defendant is guilty''.

In the Bayesian paradigm for forensic evidence reporting, it is recommended that the posterior odds, i.e.\ the LHS of~\eqref{eq:bdt}, be factored as $\text{posterior odds}=\text{prior odds}\times\text{likelihood-ratio}$~\cite{art:Gonzalez-Rodriguez2007_TASLP,phd:Ramos}. The prior odds is then moved to the RHS, where it is lumped together with the threshold as part of the legal process, reducing the scientific problem to just dealing with the likelihood-ratio. In this paper we shall eventually do so---our final results will be given in terms of likelihood-ratios. But we strongly recommend that when analysing how to compute such likelihood-ratios, one does not directly jump to working with likelihood-ratios---a habit that could easily lead to incorrect application of probability theory---see~\cite{forensic_population} and further discussion below. The safe route to deriving this kind of calculation is to first express the desired posterior and then (if possible) to factor out the likelihood-ratio. In general, a single likelihood-ratio representing the evidence cannot always be factored out---see~\cite{balding2005weight,forensic_population} for examples.

\subsection{Hypothesis posteriors}
\def\pivec{\boldsymbol{\pi}}
In what follows we shall make use of the notation of \emph{directed graphical models} (Bayesian networks)~\cite{book:Bishop_PRML} to specify conditional independence relationships. Readers are urged to familiarize themselves with this notation and how to read independence relationships from such graphs.\footnote{Observation at a node with convergent arrows (or observation at any of its descendants) induces dependency between variables linked via this node; when not observed, such nodes block
dependency. Conversely, nodes with divergent or aligned arrows induce dependency
when not observed; and block dependency when observed. A node
is `observed' if it appears to the right of the $|$ in probability notation.} 

We consider a trial that has to determine whether a defendant is the perpetrator of a crime. The prosecution hypothesis, denoted $H_1$, states that the defendant and the perpetrator are the same, while the defence hypothesis, $H_2$, states that they are different. The evidence can be separated into two parts: \emph{speech evidence}, denoted $e$, and \emph{other evidence}, denoted $\pivec$. We assume\footnote{This assumption is part of the idealized mathematical model, and is not meant to describe what happens in actual legal practice.} that the following can be computed (or is given):
\begin{align}
\label{eq:prior}
\pi_i = P(H_i|\pivec), \text{where $\pi_1+\pi_2=1$}
\end{align}
In this context we can refer to $\pivec$ as the hypothesis prior and we are not concerned about how~\eqref{eq:prior} is computed. 

What does concern us, is computation involving the speech evidence. In general, the speech evidence has two parts: speech samples known to be of the defendant and speech samples known to be of the perpetrator of the crime. In this work however, we shall eventually limit our scope to processed speech evidence in the form of automatic speaker recognizer scores.

We shall analyse our speech evidence, $e$, with the help of a generative model of the form $P(e|h,\theta)$, where $h\in\{H_1,H_2\}$ and where $\theta$ denotes the parameters of the generative model. In graphical notation, the variables we have identified so far are related as:
\begin{align}
\theta \to e \gets h \gets \pivec
\end{align}
If $\theta$ is given, the calculation of the posterior odds is well known:
\begin{align}
\label{eq:plugin}
\frac{P(H_1|e,\pivec,\theta)}{P(H_2|e,\pivec,\theta)}
&=\frac{P(H_1|\pivec)}{P(H_2|\pivec)}\times\frac{P(e|\theta,H_1)}{P(e|\theta,H_2)}
\end{align}
where the RHS is the product of prior odds and likelihood-ratio (LR). In real life however, $\theta$ is not given. If we have a suitable, large, supervised database of speech data available, $\theta$ can be estimated and plugged into~\eqref{eq:plugin} and we are done. We shall refer to this version of the likelihood-ratio as the \emph{plugin LR}:
\begin{align}
R_\text{plug}(e|\theta) &= \frac{P(e|\theta,H_1)}{P(e|\theta,H_2)}
\end{align}
It is often however the case in forensic scenarios that the large supervised databases that are available (for example LDC data) could be deemed too different from the speech encountered in the forensic case at hand. Then we may have to resort to using a \emph{small} forensic database for estimating $\theta$. Naturally, questions arise: What is the minimum size of database that I can use? What is the effect of the database size on the end-result of my calculation? 

Answers are provided by enlarging the scope of our probabilistic treatment to a so-called fully Bayesian treatment. Instead of regarding $\theta$ as given, we admit it is not given and instead assume a much weaker constraint, namely some suitable prior distribution, $P(\theta|\Pi)$, parametrized by the hyperparameters $\Pi$. We denote the (small) supervised database (or \emph{background data}) by $D=(E,L)$, where $E$ represents a collection of (processed) speech samples and $L$ represents the associated labels. The graphical notation for this more complete specification reads:
\begin{align}
\label{diag:bayes}
L \to E \gets \theta \to e \gets h \gets \pivec, 
\end{align}
with $\Pi\to\theta$ omitted for brevity. This diagram makes explicit the assumption that the background data, $E$, and the evidence, $e$, were generated by the very same model parameters. Taking into account the independence assumptions encoded by~\eqref{diag:bayes}, the posterior odds can be expressed as:
\begin{align}
\label{eq:bayes}
\frac{P(H_1|e,D,\pivec,\Pi)}{P(H_2|e,D,\pivec,\Pi)}
&=\frac{P(H_1|\pivec)}{P(H_2|\pivec)}\times\frac{P(e|D,\Pi,H_1)}{P(e|D,\Pi,H_2)}
\end{align}
This Bayesian posterior odds also factors into prior odds and a \emph{Bayesian LR}, which we denote:
\begin{align}
\label{eq:lr}
R_B(e|D) &= \frac{P(e|D,\Pi,H_1)}{P(e|D,\Pi,H_2)}
\end{align}

\subsection{The Bayesian likelihood-ratio}
\def\Ex#1#2{\big\langle#1\big\rangle_{#2}}
\def\EX#1#2{\Big\langle#1\Big\rangle_{#2}}
To compute the LR~\eqref{eq:lr}, we can express the numerator or denominator as:
\begin{align}
\label{eq:pred}
\begin{split}
P(e|D,\Pi,H_i) &= \int_\Theta P(e,\theta|D,\Pi,H_i) \,d\theta \\
&= \int_\Theta P(e|\theta,D,\Pi,H_i)P(\theta|D,\Pi,H_i) \,d\theta \\
&= \int_\Theta P(e|\theta,H_i)P(\theta|D,\Pi) \,d\theta 
\end{split}
\end{align}
where $\Theta$ represents the support of $P(\theta|\Pi)$ and where we have dropped irrelevant conditioning terms using the independence relationships encoded by~\eqref{diag:bayes}. We shall refer to $P(e|\theta,H_i)$ as the \emph{plugin likelihood} and to $P(\theta|D,\Pi)$ as the \emph{parameter posterior}. The parameter posterior represents everything we know about the parameters after having processed the background data $D$. The required LR now becomes: 
\begin{align}
\label{eq:blr}
\begin{split}
R_B(e|D) &= \frac{\int_\Theta P(e|\theta,H_1)P(\theta|D,\Pi) \,d\theta}{\int_\Theta P(e|\theta,H_2)P(\theta|D,\Pi) \,d\theta} \\
&= \frac{\Ex{P(e|\theta,H_1)}{P(\theta|D,\Pi)}}{\Ex{P(e|\theta,H_2)}{P(\theta|D,\Pi)}}
\end{split}
\end{align}
which is a \emph{ratio of expected likelihoods}, with expectations taken w.r.t.\ the parameter posterior. 

If however, one immediately starts with the agenda of \emph{let's calculate the likelihood-ratio}, without referring to a careful specification as in~\eqref{diag:bayes}, it is easier to fall into the trap of doing \emph{erroneous} calculations like taking expectations of the plugin LR or its logarithm:
\begin{align*}
\EX{\frac{P(e|\theta,H_1)}{P(e|\theta,H_2)}}{\theta} \; &\text{ or }\; \EX{\log\frac{P(e|\theta,H_1)}{P(e|\theta,H_2)}}{\theta} 
\end{align*}  
In fact, following our derivation in~\cite{NB_ICASSP14}, we can make the difference between the expected plugin log-LR and the correct Bayesian LR explicit:
\begin{align}
\begin{split}
\label{eq:b_vs_p}
\log R_B(e|D) &= \Ex{\log R_\text{plug}(e|\theta)}{\theta}\\ 
&+ \EX{\log\frac{P(\theta|D,e,H_2,\Pi)}{P(\theta|D,e,H_1,\Pi)}}{\theta}
\end{split}
\end{align}
where $P(\theta|D,e,H_i,\Pi)$ is an \emph{augmented parameter posterior} conditioned on $D$ as well as one additional supervised example, $e$, with assumed label $H_i$. This relationship holds for expectations over any distribution for $\theta$ that avoids zero denominators.

Consider the following scenario. A forensic practitioner is tasked with summarizing the weight of the speech evidence, $e$, for use in court. He decides to provide this via $R_\text{plug}(e|D)$. He realizes however, that his $D$ is small and that if he had happened to have had some other small database, say $D'$, the value $R_\text{plug}(e|D')$ could have been substantially different. He therefore goes and simulates the selection of many similarly sized databases and for each such simulation computes $\log R_\text{plug}(e|D')$, keeping $e$ constant, but varying $D'$. Armed with a collection of such values, he can now summarize his findings as $\log(\text{LR})=\mu\pm\sigma$, where $\mu$ and $\sigma$ are the mean and standard deviation of the collection of simulated values. 

What is the court to do with $\mu$ and $\sigma$? How does this help the court to decide between $H_1$ and $H_2$? The court could perhaps decide, if $\sigma$ is too large, that the evidence cannot be trusted and that the decision should be based solely on $\frac{P(H_1|\pi)}{P(H_2|\pi)}$. Or, the court could decide, if $\sigma$ is small enough, to go ahead and use $\mu$ to represent the weight of evidence. But $\mu=\Ex{\log R_\text{plug}(e|\theta)}{\theta}$ where the distribution for this expectation is formed by the above sampling process. Our~\eqref{eq:b_vs_p} shows that $\mu$ cannot  give the correct value (except perhaps by accident), because the second term is ignored. 

More generally, any procedure that attempts to represent the evidence via a probability distribution over the plugin LR (e.g.\ \cite{LR_interval}) fails to enable the court to compute the posterior probabilities required by~\eqref{eq:bdt}. Submitting LR distributions to court is therefore contrary to Balding's principle that ``evidence is of value inasmuch as it alters the probability that the defendant is guilty''. 

If we want to work within the constraints of Bayes decision theory, probability theory should be followed to compute the posteriors. In the case of the simple model in~\eqref{diag:bayes}, the solution is given by~\eqref{eq:blr}.

\subsection{Integrating out $\theta$}
Computing $R_B$ via~\eqref{eq:blr} requires the computation of integrals w.r.t.\ $\theta$. For a restricted class of conjugate-exponential models this can be done in closed form (we show an example later), but in general, closed-form solutions are not available~\cite{book:Bishop_PRML}. This means the integrals have to be approximated. This section provides some general advice to help to avoid gross inaccuracy with such approximation.

The first apparent problem is that the parameter posterior, $P(\theta|D,\Pi)$, which is required for~\eqref{eq:blr}, conceals a similar integral. Recalling $D=(E,L)$ and using~\eqref{diag:bayes}, we have:
\begin{align}
P(\theta|D,\Pi) &= P(\theta|E,L,\Pi) = \frac{P(\theta|\Pi)P(E|L,\theta)}{P(E|L,\Pi)}
\end{align}
where the \emph{normalizer} is:
\begin{align}
P(E|L,\Pi) &= \int_\Theta P(\theta|\Pi)P(E|L,\theta)\,d\theta
\end{align}
But, since this normalizer is independent of $h$ and $\theta$, it cancels in~\eqref{eq:blr}, giving:
\begin{align}
\begin{split}
\label{eq:blr2}
R_B(e|D) &= \frac{\int_\Theta P(e|\theta,H_1)P(E|\theta,L) P(\theta|\Pi)\,d\theta}{\int_\Theta P(e|\theta,H_2)P(E|\theta,L) P(\theta|\Pi)\,d\theta} \\
&=\frac{\int_\Theta P(e,E|\theta,H_1,L) P(\theta|\Pi)\,d\theta}{\int_\Theta P(e,E|\theta,H_2,L)P(\theta|\Pi)\,d\theta} \\
&=\frac{P(e,E|H_1,L,\Pi)}{P(e,E|H_2,L,\Pi)}
\end{split}
\end{align}
where the new numerator and denominator are just the normalizers of our previously introduced augmented parameter posteriors, $P(\theta|D,e,H_i,\Pi)$. 

We see now that the real challenge with the calculation of $R_B$ is not the normalizer for $P(\theta|D,\Pi)$, but rather calculation of two separate normalizers for each $P(\theta|D,e,H_i,\Pi)$.

A common practice for approximation in Bayesian calculations is to obtain an approximate parameter posterior, say $\tilde P(\theta|D,\Pi)$, on which further calculations are based. In our case that would give:
\begin{align}
\label{eq:badapprox}
R_B(e|D) &\approx \frac{\int_\Theta P(e|\theta,H_1)\tilde P(\theta|D,\Pi) \,d\theta}{\int_\Theta P(e|\theta,H_2)\tilde P(\theta|D,\Pi) \,d\theta}
\end{align}
Usually $\tilde P(\theta|D,\Pi)$ is chosen to be a good approximation of the true posterior near its peak, while the tails receive little attention. Unfortunately, one or both of $P(e|\theta,H_i)$, as a function of $\theta$, could have its peak far from the accurate region of $\tilde P$ and could therefore effectively be sampling the approximate posterior in inaccurate regions. In this situation, using independent approximations for the numerator and denominator of~\eqref{eq:blr2} could be more accurate.    

However, \eqref{eq:blr2} is not without its own pitfalls. In a situation where $E$ contains very many examples, we could have $P(E|\theta,L) \gg P(e|\theta,H_i)$ and then finite numerical precision could cause the numerator and denominator representations to become identical. Care should be taken to decide which approximation is best under the circumstances.

\section{Bayesian calibration}
\def\ND{\mathcal{N}}
\def\GD{\mathcal{G}}
Unfortunately, current generative models for speaker recognition, such as PLDA~\cite{KennyOdyssey10}, are too complex~\cite{jesus_interspeech11} and perhaps at the same time still too inaccurate, to allow direct Bayesian calculation of the kind required here. That is, if we represent defendant and perpetrator speech samples as i-vectors~\cite{Dehak10frontend}, so that $e$ is a pair of i-vectors, and use PLDA as the generative model, then an accurate calculation of $R_B$ is intractable. In fact, PLDA does not even give accurate plugin likelihood-ratios. To obtain well-calibrated likelihood-ratios from a speaker recognizer, an intermediate step, known as \emph{calibration}~\cite{art:brummer_csl_2006,PhD} is required---for recent examples see~\cite{miranti,DvL_IS13,NB_IS13,NB_ICASSP14,NB_Odyssey14,SRI_odyssey04}. 

For our current purposes, calibration can be understood as follows. We process the original speech evidence via an automatic speaker recognizer which processes the defendant and perpetrator speech samples to output a real score. A high (more positive) score favours $H_1$, while a low (more negative) score favours $H_2$. This score is now considered to be the evidence, $e$ for the trial at hand. Our supervised database, $D=(E,L)$, has $E=s_1,\ldots,s_n$, a collection of $n$ scores generated by the same recognizer in response to $n$ pairs of speech samples deemed similar the ones in the trial at hand. The labels, $L=\ell_1,\ldots,\ell_n\in\{H_1,H_2\}$, indicate whether each score was computed from samples satisfying $H_1$ or $H_2$. In summary, $e,h$ refer to the score and hypothesis of the trial at hand, while the $s_t,\ell_t$ refer to scores and hypotheses of the background data.

\subsection{Normal score model, with conjugate prior}
Bayesian calibration presents challenges in addition to the above-mentioned integration problem. 

The problem in the plugin method, of choosing a family of score distributions~\cite{NB_Odyssey14}, is shared by the Bayesian method. For this paper we default to Gaussian score modelling, which will suffice to demonstrate the advantages of Bayesian calibration. Future work will examine more sophisticated score models.

The challenge of assigning the prior, $P(\theta|\Pi)$, should not be underestimated. According to Jaynes~\cite{book:Jaynes_PTTLOS_2003}, priors represent the unfinished half of probability theory. For some thoughts on selecting priors for forensic problems, see~\cite{dnatut}. Here we aim for simplicity and select a conjugate prior, which gives closed-form integrals.

We let our model parameters be $\theta=(\mu_1,\mu_2,\lambda_1,\lambda_2)$, for a Gaussian model of the form:
\begin{align}
P(e|H_i,\theta) &= \ND(e|\mu_i,\lambda_i^{-1})
\end{align}
where the $\mu_i$ are means and the $\lambda_i$, are precisions (inverse variances). We consider the background data to be generated iid from $\theta$, so that:
\begin{align}
P(E|L,\theta) = \prod_{i=1}^2\; \prod_{t:\ell_t=H_i} \ND(s_t|\mu_i,\lambda_i^{-1})
\end{align}
The conjugate prior is Gaussian-gammma, of the form~\cite{book:Bishop_PRML}:
\begin{align}
P(\theta|\mu_0,\beta,a,b) &= \prod_{i=1}^2\ND(\mu_i|\mu_0,\beta^{-1}\lambda^{-1})\GD(\lambda_i|a,b)
\end{align}
where $\GD$ is a gamma distribution with parameters $a,b>0$. By choosing $a=b\ll1$, $\lambda_i$ has an expected value of $1$ and a very large variance, making this prior non-informative about the scale of the scores. Likewise, to be non-informative about the score location, we can arbitrarily choose $\mu_0=0$ and $\beta\ll1$, which gives very large variance for $\mu_i$. 

Finally, $R_B(e|D)$ is computed by solving~\eqref{eq:pred} in closed form, which results in a T distribution---for details, see~\cite{fbgc,minka_gaussian,book:Bishop_PRML}.

\begin{figure}[!htb]
\centerline{\includegraphics[width=0.5\textwidth,trim = 1.8cm 1.3cm 1.8cm 1.6cm,clip = true]{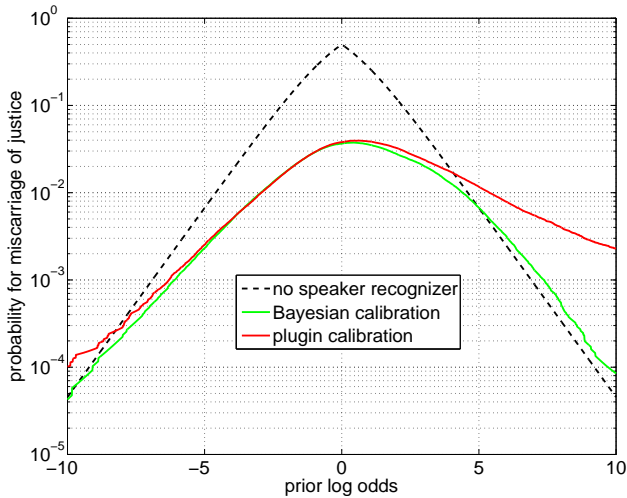}}
\caption{Bayesian vs Plugin calibration, trained on 9 $H_1$ and 27 $H_2$ examples.}  
\label{fig:calbration1}
\end{figure}

\begin{figure}[!htb]
\centerline{\includegraphics[width=0.5\textwidth,trim = 1.8cm 1.3cm 1.8cm 1.6cm,clip = true]{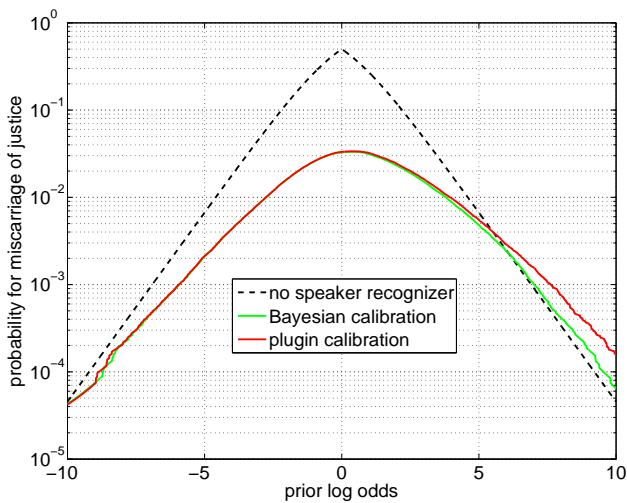}}
\caption{Bayesian vs Plugin calibration, trained on 30 $H_1$ and 405 $H_2$ examples.}  
\label{fig:calibration2}
\end{figure}

\begin{figure}[!htb]
\centerline{\includegraphics[width=0.5\textwidth,trim = 1.8cm 1.3cm 1.8cm 1.6cm,clip = true]{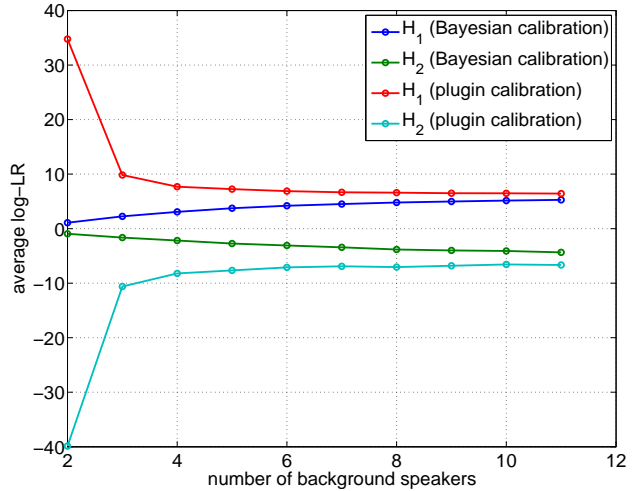}}
\caption{Plugin calibration becomes overconfident with insufficient data. Bayesian calibration moderates its confidence as a function of the data size.}  
\label{fig:calibration3}
\end{figure}

\section{Experiments}
Our experimental setup is similar to that in~\cite{NB_ICASSP14,NB_Odyssey14}, using scores from a single speaker recognizer (an i-vector PLDA system), which was part of the ABC submission~\cite{ABC12} to the NIST SRE'12 speaker recognition evaluation~\cite{web:sre12}. We selected background scores, $D$, from the ABC development database, containing pre-SRE'12 speech, while the scores for the evidence in the trial at hand, $e$, came from SRE'12.

We constructed $D$ using either 3 or 10 speakers, with 3 recordings per speaker. For 3 speakers, this gave 9 $H_1$ scores and 27 $H_2$ scores, while for the 10 speaker case this gave 30 $H_1$ and $405$ $H_2$ scores. The error-rates were averaged in each case over 1000 different random selections of speakers for $D$.

To test how well it works, we vary the hypothesis prior $\pi_1$ over a wide range of values and plot the error-rate that results when using our calibrated likelihood-ratios to make decisions using~\eqref{eq:bdt}. In a real court case the relative costs matter, but for our purposes of evaluating accuracy, we can conveniently set the costs to 1 without loss of generality~\cite{PhD,BOSARIS}. This error-rate can be interpreted as the probability for a miscarriage of justice (false acquit, or false convict). 

The results are shown for $D$ having 3 speakers (figure~\ref{fig:calbration1}) and for 10 speakers (figure~\ref{fig:calibration2}), where we compare: (i) traditional plugin calibration; (ii) the proposed Bayesian calibration; and (iii) simply convicting if $\pi_1>\pi_2$, giving an error-rate of $\min(\pi_1,\pi_2)$. The horizontal axis is $\log\frac{\pi_1}{1-\pi_1}=\log\frac{\pi_1}{\pi_2}$.

Figure~\ref{fig:calibration3} compares the average hypothesis-conditional log-LR values for plugin vs Bayesian, as a function of the amount of background data. The Bayesian method behaves intuitively---and the plugin method counterintuitively. 

\subsection{Discussion}
Both plugin and Bayesian calibrations face two main challenges on this data. The Bayesian method solves the first challenge:

(i) Limited training data. When the data is very limited (3 speakers) the Bayesian method clearly does better. With more data (10 speakers) the problem is less severe and the gap closes. 

(ii) This data exhibits mild \emph{dataset shift}~\cite{book:datasetshift}, where the score distributions change between background and test sets. Neither method has a mechanism for dealing with dataset shift. Future work should examine ways of including dataset shift modelling into the Bayesian method.    

\section{Conclusion}
We have shown that by widening the scope of our probabilistic treatment, the applicability of score calibration can be enlarged to include challenging situations with very limited background data. With limited data, the traditional plugin method misbehaves by being overconfident, while the Bayesian method moderates its confidence as a function of the amount of data.

\newpage
\eightpt
\bibliographystyle{IEEEtran}
\bibliography{refs}
\end{document}